\documentclass[conference]{IEEEtran}
\IEEEoverridecommandlockouts
\usepackage{cite}
\usepackage{amsmath,amssymb,amsfonts}
\usepackage{algorithmic}
\usepackage{graphicx}
\usepackage{textcomp}
\usepackage{amsthm}
\usepackage{xcolor}
\usepackage{bbding}
\usepackage{makecell}
\usepackage{algorithm}
\usepackage{color}
\usepackage{multirow}
\usepackage{bm}
\usepackage{array}
\usepackage{booktabs}
\usepackage{tabularx,booktabs}
\usepackage{multicol}
\usepackage{cleveref}
\usepackage{balance}
\usepackage{lastpage}
\usepackage{tabulary}
\usepackage{subfigure}
\usepackage{etoolbox}
\usepackage{bbm}
\usepackage{enumitem}
\usepackage{mathrsfs}
\usepackage{multicol}
\usepackage{amsfonts,amssymb}
\usepackage{epstopdf}
\usepackage{ulem}
\usepackage{cancel}
\usepackage{setspace}
\usepackage{stackengine}
\usepackage{tcolorbox}
\tcbuselibrary{breakable}
\usepackage{threeparttable}

\usepackage{url}
\usepackage{amsthm}
\theoremstyle{definition}

\renewenvironment{IEEEbiography}[1]
  {\IEEEbiographynophoto{#1}}
  {\endIEEEbiographynophoto}

\usepackage{nomencl}
\makenomenclature

\def\BibTeX{{\rm B\kern-.05em{\sc i\kern-.025em b}\kern-.08em
    T\kern-.1667em\lower.7ex\hbox{E}\kern-.125emX}}

\newcommand{\blue}[1]{\textcolor{black}{#1}}

\makeatletter
\def\ps@IEEEtitlepagestyle{%
  \def\@oddhead{%
    \vbox{%
      \hbox to\textwidth{\hfil This work has been accepted by IEEE Wireless Communications Magazine\hfil}%
      \vskip 3pt
      \hrule height 0.4pt
    }%
  }%
  \let\@evenhead\@oddhead
  \let\@oddfoot\@empty
  \let\@evenfoot\@empty
}
\makeatother

\begin{document}

\bstctlcite{IEEEexample:BSTcontrol}

\title{Large Language Models (LLMs) for Telecom Root Cause Analysis (RCA): A Structured Reasoning Framework for Evidence-Grounded Diagnosis \thanks{Hao Zhou is with Samsung Research America, Toronto, Ontario, ON M5E0C5.\ (email: haozhou029@gmail.com); Mandar Kulkarni and Yan Xin are with Samsung Research America, Berkeley Heights, NJ 07922
USA  (email: \{m.kulkarni2, yan.xin\}@samsung.com);
Hao Chen and Charlie (Jianzhong) Zhang are with Samsung Research America, Plano, Texas, TX 75023, USA. (email: \{hao.chen1, jianzhong.z\}@samsung.com);}}

\author{\IEEEauthorblockN{Hao Zhou, Mandar Kulkarni, Hao Chen, Yan Xin, and Charlie (Jianzhong) Zhang,  \IEEEmembership{Fellow, IEEE}}}

\maketitle
\begin{abstract}
Root cause analysis (RCA) is a critical task in telecom network operations, but diagnosing performance degradations in modern 5G and emerging 6G networks remains challenging due to complex cross-layer dependencies. While large language models (LLMs) offer promising capabilities for reasoning and knowledge integration, directly applying vanilla LLMs to telecom RCA often leads to hallucination, unstable reasoning, and poor alignment with structured network evidence.
This work first reviews the evolution of telecom RCA from rule-based and machine learning (ML) approaches to emerging LLM-enabled techniques, and provides an overview of recent paradigms, including structured reasoning, retrieval-augmented knowledge grounding, agentic orchestration, and verifiable reasoning.
Building upon these insights, we propose a structured reasoning framework for LLM-enabled telecom RCA that aligns diagnostic reasoning with telecom-specific evidence and domain knowledge.
The proposed approach first organizes heterogeneous network telemetry into canonical contexts, and then enforces decision-path reasoning during diagnosis, and finally generates evidence-grounded explanations for reliable fault identification.
Experimental results on two 5G RCA datasets, TeleLogs and TelecomTS, demonstrate that the proposed framework consistently improves diagnostic accuracy and decision consistency compared with baseline techniques.
These cross-dataset results highlight the importance of structured reasoning design for practical LLM-based RCA systems in next-generation telecom networks.
\end{abstract}


\section{Introduction}

With the deployment of 5G-Advanced and the emerging 6G networks, telecom systems are evolving into highly complex and large-scale infrastructures.
These networks integrate heterogeneous components across radio, transport, and core layers, making reliable operation increasingly challenging.
As a result, network failures and service degradations may lead to significant operational consequences. For example, the Rogers outage in Canada (2022), the AT\&T outage in the United States (2024), and the Optus outage in Australia (2023) disrupted connectivity for over 100 million devices, blocked more than 25,000 emergency calls, and caused nationwide disruptions to payment systems, transportation networks, and healthcare services, with more than \$100 million loss~\cite{rogers2022outage}.

Traditionally, telecom RCA has relied on rule-based expert systems and, more recently, ML techniques~\cite{murphy2023fault}. In particular, early RCA solutions rely on predefined alarm patterns or KPI thresholds to trigger corresponding troubleshooting procedures.
%
To improve scalability, data-driven ML models have been introduced to detect and classify potential faults. These approaches can process large volumes of telemetry data, but they primarily perform statistical pattern recognition rather than explicit diagnostic reasoning. As a result, ML-based RCA systems often struggle to adapt to unseen failure modes without extensive retraining~\cite{qiu2023ai}.

\begin{figure*}[t]
  \centering
  \includegraphics[width=1\textwidth]{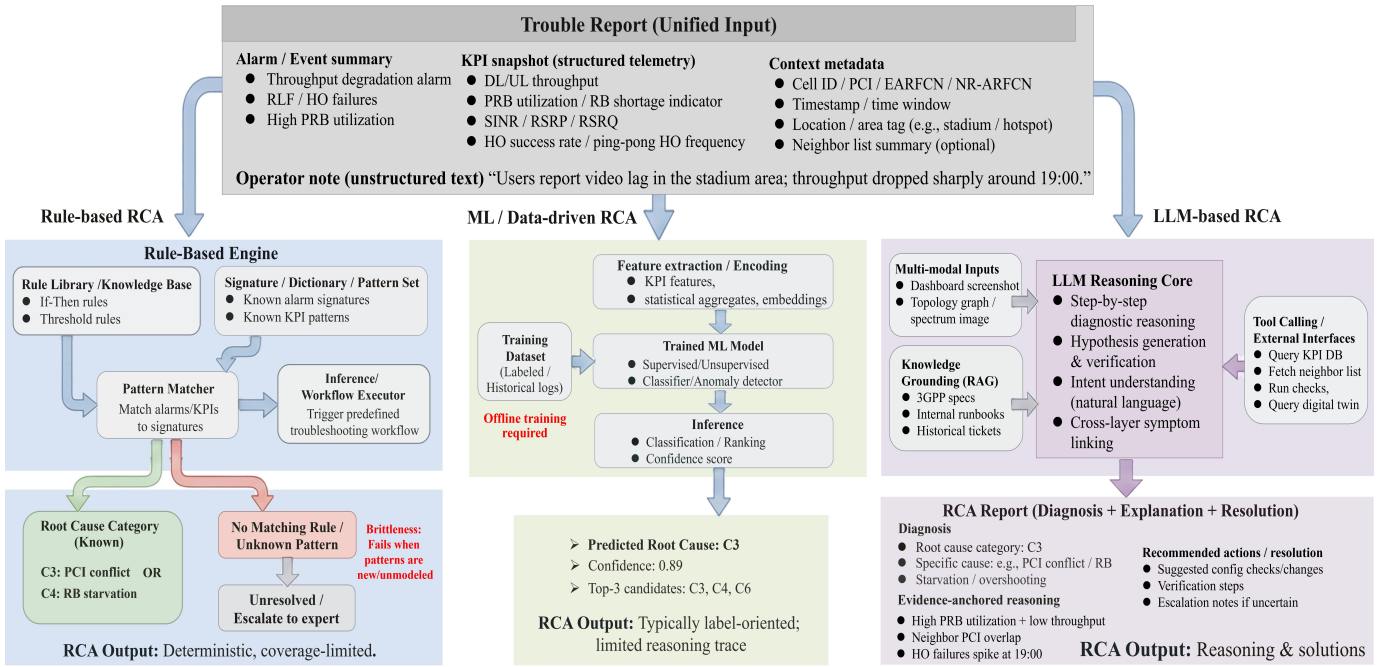}
  \caption{Comparison of rule-based, ML, and LLM-enabled RCA approaches.}
  \label{fig:compare}
\end{figure*}

Recently, large language models (LLMs) have been successfully applied to many fields, and they also offer promising opportunities for telecom, e.g., knowledge understanding~\cite{yuan2025understanding}, power control~\cite{zhou2025prompting}, traffic prediction~\cite{hu2025self}, network management~\cite{zhou2024large}, etc.
LLMs have several critical advantages over traditional rule-based and ML approaches for telecom RCA. Firstly, LLMs possess strong reasoning capabilities that enable multi-step diagnostic analysis. Unlike rule engines that follow predefined logic and ML models that primarily learn statistical correlations, LLMs can perform step-by-step reasoning across multiple system indicators to infer potential root causes. 
Secondly, LLMs can integrate knowledge across heterogeneous domains, e.g., radio access, transport, and core networks. Such cross-domain reasoning is difficult for traditional methods that typically operate on isolated feature sets.
Thirdly, LLMs can generate human-readable explanations of their diagnostic process. This capability improves the interpretability of RCA outcomes and helps network engineers understand how specific evidence leads to a particular RCA diagnosis.

However, despite these advantages, directly applying vanilla LLMs to telecom RCA remains challenging. Firstly, the reasoning process of LLMs can be unstable. Specifically, telecom RCA often requires systematic verification of multiple diagnostic factors, such as mobility behavior, resource scheduling, and coverage geometry. Vanilla LLMs may produce inconsistent reasoning paths when interpreting similar sets of network measurements~\cite{sana2025reasoning}.
Secondly, telecom RCA involves heterogeneous and structured evidence from multiple system layers. Without explicit mechanisms to align structured telemetry with diagnostic reasoning, LLMs may rely on superficial correlations rather than causal relationships between network events and performance degradations~\cite{wu2025tn}.
Thirdly, LLMs are prone to hallucination. For instance, in telecom RCA scenarios, incorrect conclusions may arise when the model speculates about potential causes without grounding its reasoning in concrete indicators such as KPIs, topology relationships, or configuration parameters~\cite{zou2025telecomgpt}.

To address these challenges, this work proposes a structured evidence- and knowledge-aligned fine-tuning framework (SEKA-FT) for evidence-grounded telecom RCA.
Rather than treating RCA as an input-to-label prediction task, we formulate it as a structured reasoning problem, where heterogeneous network evidence must be progressively examined before a final root-cause decision is made.
Under this formulation, the diagnostic process is explicitly guided by an evidence-to-path-to-decision structure.
First, heterogeneous troubleshooting inputs are reorganized into canonical context blocks so that user-plane KPIs, mobility statistics, neighbor-cell relations, and engineering parameters can be interpreted consistently.
Second, decision-path reasoning is introduced to decompose RCA into staged diagnostic checks, enabling the model to rule out unsupported hypotheses before producing the final answer.
Third, evidence-grounded explanations are incorporated into the supervision target so that the generated diagnoses remain traceable to concrete telecom indicators and domain knowledge.
Therefore, SEKA-FT jointly aligns structured evidence, intermediate diagnostic paths, final RCA labels, and traceable explanations within a unified fine-tuning process.
Through this design, telecom RCA is transformed from prediction-based troubleshooting into decision-path controlled reasoning. This enables the model to make RCA decisions based not only on final label supervision, but also on structured diagnostic evidence and intermediate reasoning consistency. SEKA-FT improves both the accuracy of root-cause identification and the trustworthiness of generated diagnosis.

The contributions of this work are two-fold.
First, we provide a systematic overview of the evolution of RCA in telecom networks and analyze the emerging role of LLMs in next-generation RCA systems.
In particular, we discuss how LLM-enabled paradigms can enhance RCA capabilities through structured reasoning, retrieval-augmented generation (RAG), agentic orchestration, and reinforcement learning with verifiable rewards (RLVR).
Second, we develop SEKA-FT as a telecom-RCA-specific fine-tuning framework that \textcolor{black}{integrates heterogeneous evidence, intermediate diagnostic checks, root-cause decisions, and evidence-grounded explanations into a unified evidence-to-path-to-decision supervision structure}.
\textcolor{black}{The technical contribution lies in jointly encoding these elements according to the diagnostic dependencies of telecom RCA, such that the model learns not only the final label but also a structured reasoning path consistent with telecom evidence and domain constraints.}
Extensive case studies on two 5G RCA datasets, TeleLogs and TelecomTS, demonstrate consistent improvements over ICL, non-LLM, and SFT-based baselines.
\blue{These results show that the proposed framework is not limited to a single dataset template, but remains effective across distinct 5G RCA and observability settings with different evidence representations and diagnostic characteristics.}

\section{RCA: From Rule-based and ML to LLMs}

As shown in Fig.~\ref{fig:compare}, the evolution of RCA can be viewed as a shift from deterministic rule matching, to statistical label prediction, and further toward evidence-grounded reasoning.
First, rule-based expert systems provide transparent and deterministic behavior, but they rely on predefined alarm patterns, KPI thresholds, and manually maintained troubleshooting logic. As network scenarios become more diverse, such static rules are difficult to scale and update.

Data-driven ML methods improve scalability by learning statistical patterns from historical telemetry. They can detect anomalies, classify faults, and rank possible root causes from large volumes of KPIs, alarms, and logs.
However, most ML-based RCA methods still operate as input-to-label mapping systems. They usually output a probability distribution or a predicted class, but do not explicitly describe how different pieces of evidence support candidate root causes. Their performance also depends strongly on training-data coverage and distribution stability.

LLMs introduce a new opportunity for telecom RCA because they can combine structured telemetry, textual descriptions, engineering knowledge, and operational instructions within a unified reasoning interface.
Compared with conventional classifiers, LLMs can generate human-readable diagnostic traces and support multi-step analysis across heterogeneous evidence sources.
However, the presence of language-level reasoning ability does not automatically make LLMs reliable RCA engines. Without structured evidence alignment, an LLM may produce plausible but unsupported explanations, follow unstable reasoning paths, or rely on superficial correlations between input fragments and root-cause labels.

Therefore, the key challenge is not merely to apply an LLM to telecom RCA, but to control how the model connects heterogeneous evidence to diagnostic decisions.
This motivates the structured reasoning design adopted in this work: RCA should be formulated as an evidence-grounded reasoning process in which intermediate diagnostic checks, final labels, and explanations are jointly aligned.

\begin{figure*}[t]
  \centering
  \includegraphics[width=0.9\textwidth]{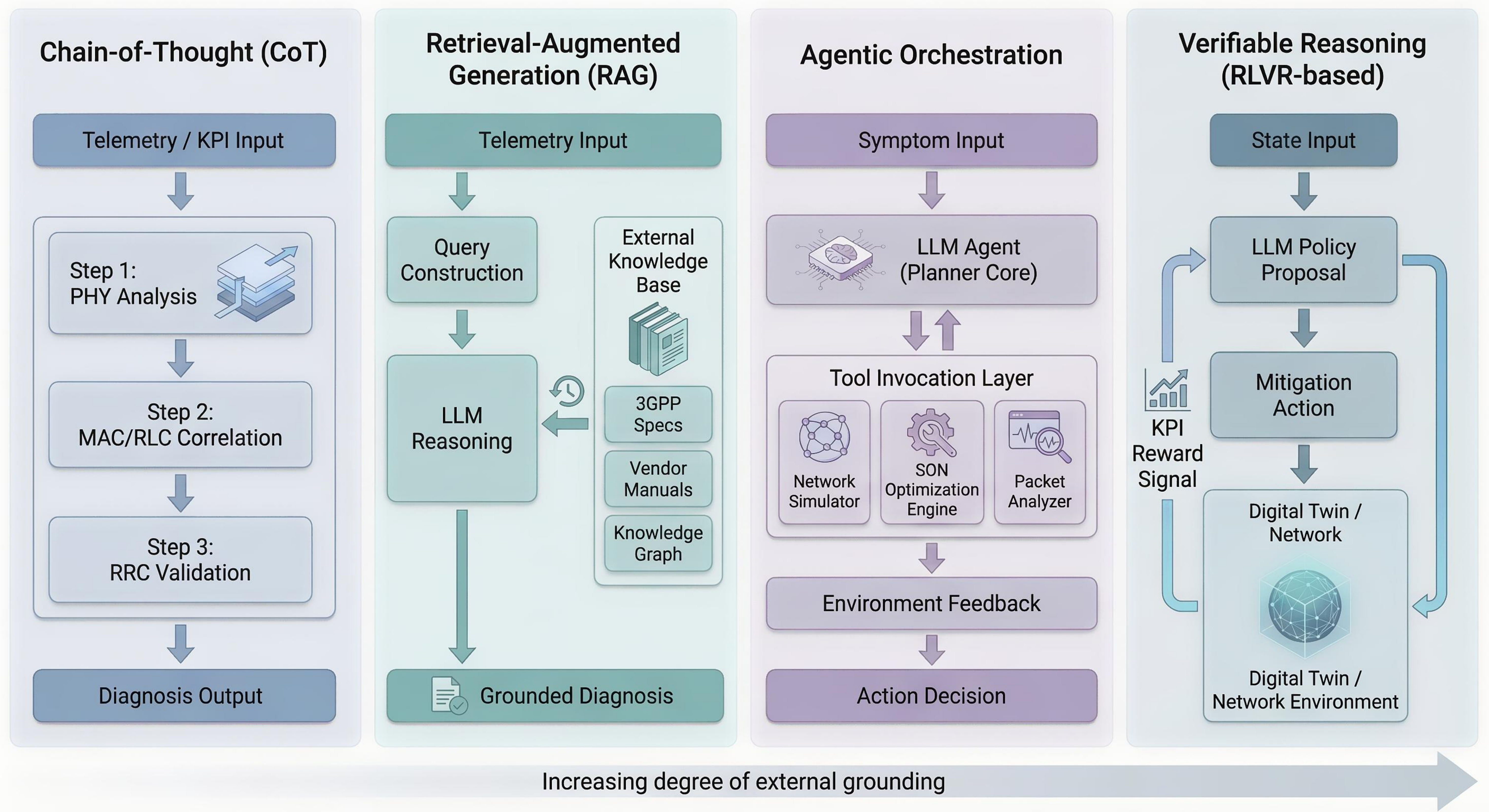} 
  \caption{Summary of various LLM-enabled techniques for telecom RCA. The arrow at the bottom illustrates a progressive expansion of reasoning capability. From left to right, the LLM evolves from internally structured reasoning (CoT), to externally grounded knowledge integration (RAG), to active environmental interaction (Agentic orchestration), and ultimately to physically verifiable alignment through outcome-based feedback (RLVR). This progression reflects an increasing degree of external grounding and real-world coupling in telecom RCA.}
  \label{fig:summary}
\end{figure*}

\section{LLM-enabled Techniques for Telecom Root Cause Analysis}

This section presents a progression path of LLM-enabled paradigms for telecom RCA.

\subsection{Chain-of-Thought: Structuring Diagnostic Logic}

CoT is a prompting technique that encourages LLMs to generate a series of intermediate reasoning steps before reaching a final answer~\cite{zhou2025large22}.
In telecom RCA, this is important because one-shot inference often fails to capture cross-layer dependencies. For example, when a gNodeB reports downlink throughput degradation, a vanilla LLM may directly attribute the issue to congestion without checking control-plane events.
A structured CoT can instead guide the model through staged verification: first evaluating physical-layer indicators, then checking MAC/RLC-layer evidence, and finally correlating the observations with RRC or handover-related signalling logs.
This illustrates why RCA should be organized as an explicit diagnostic path rather than a direct label prediction process.

\subsection{Retrieval-Augmented Generation: Knowledge Grounding}

Standard LLMs often produce plausible but technically incorrect hallucinations.
To this end, RAG is a framework that allows an LLM to retrieve relevant information from an external, authoritative knowledge base before generating a response~\cite{yuan2026enhancing}.
For telecom RCA, a RAG system can retrieve relevant materials such as 3GPP specifications, vendor manuals, historical incident reports, or verified troubleshooting rules based on the current telemetry or alarm context.
The retrieved evidence can then constrain the LLM's diagnosis and reduce hallucination.
For instance, when throughput degradation is observed in a serving cell, the system may trace related transport, core-network, or neighboring-cell dependencies that are not directly visible from local KPIs.
This shows that reliable LLM-based RCA requires not only language reasoning, but also external grounding in domain knowledge and network structure.

\subsection{Agentic Orchestration: Environment-interactive Planning}

An agentic LLM system differs from a standard chatbot in its ability to use tools and execute multi-step plans autonomously to achieve a goal.
While a direct LLM interface is limited to the information provided in the prompt, an LLM agent can interact with external systems through API calls~\cite{shi2025leveraging}.
In telecom RCA, an agentic framework can treat the LLM as a coordinator that verifies diagnostic hypotheses using specialized tools. For example, the agent may trigger an ns-3 simulation to estimate the impact of a configuration change, or query a self-organizing network module for antenna-tilt or power-control recommendations.
Furthermore, the agentic paradigm can naturally extend to a multi-agent system, where the complex RCA workflow is decomposed into specialized roles, forming a collaborative “expert team”.
Different agents may focus on solution planning, log analysis, configuration validation, etc. Such orchestration is useful for multi-domain failures where no single evidence source is sufficient.

\subsection{RLVR: Physically Verifiable Reasoning and Alignment}

Verifiable reasoning refers to a paradigm where the model's output is not just judged by its linguistic fluency, but by its ability to satisfy checkable conditions or rewards. Standard LLMs are typically aligned using human preference, which is subjective and may not reflect technical truth~\cite{wen2025reinforcement}.
In the RLVR paradigm, a network environment such as a high-fidelity simulator or digital twin can serve as a feedback source. After the LLM proposes a diagnostic path and mitigation action, the system evaluates whether target KPIs such as throughput, packet loss, or handover failure rate improve.
Successful reasoning trajectories can be reinforced, while ineffective ones can be penalized. This shifts the alignment objective from human preference alone to operational effectiveness under network constraints.

In summary, LLM-enabled RCA does not rely on a single technique, but evolves through progressively stronger forms of grounding and environmental coupling.
CoT emphasizes diagnostic path organization, RAG provides knowledge grounding, agentic orchestration enables tool-based hypothesis verification, and RLVR introduces outcome-based alignment.
These insights motivate the proposed SEKA-FT framework, which focuses on structured evidence alignment and decision-path control as a practical step toward reliable LLM-based RCA.

\section{SEKA-FT: Structured Evidence- and Knowledge-Aligned Fine-Tuning for RCA Reasoning}

The previous sections discussed the evolution of LLM-enabled RCA paradigms, ranging from structured reasoning to verifiable RLVR. However, additional domain-specific challenges may prevent real-world RCA applications.
This section introduces SEKA-FT, a structured fine-tuning framework for telecom RCA decision-path control.
\textcolor{black}{At its core, SEKA-FT formulates telecom RCA fine-tuning around a unified evidence-to-path-to-decision supervision structure, instantiated through three tightly coupled components: canonical context structuring for stable RCA fine-tuning; CoT-enabled decision-path control for reliable RCA reasoning; evidence- and knowledge-anchored explanation design for hallucination mitigation.}
The details are introduced below.

\begin{figure*}[t]
\centering
\begin{tcolorbox}[title = { SEKA-FT input/output sample (ID\_92TVA88GDA)}]
\small
\{``id'': ``ID\_92TVA88GDA'', \\
``input\_text'': ``Analyze the 5G wireless network drive-test user plane data and engineering parameters. Identify the reason for the throughput dropping below 600Mbps in certain road sections. From the following 7 potential root causes, select the most likely one.\\
M1: Average scheduled RBs are below 160, affecting throughput. \\
M2: Frequent handovers degrade performance.\\
M3: The serving cell's downtilt angle is too large, causing weak coverage at the far end.\\
M4: Test vehicle speed exceeds 40 km/h, impacting user throughput.\\
M5: Neighbor cell and serving cell have the same PCI mod 30, leading to interference.\\
M6: The serving cell's coverage distance exceeds 1 km, resulting in over-shooting.\\
M7: A neighboring cell provides higher throughput.

\textbf{Selected Serving Cell (Reasoning Anchor)}: Serving Cell PCI: 0
Neighbor Cell PCI at bottleneck: 20, 4\\
\textbf{Bottleneck Snapshot}:
Downlink throughput: 420.88 Mbps.
Serving SS-RSRP: -94.67 dBm.
Serving SS-SINR: 8.43 dB.
Scheduled RBs: 177.0.
Serving gNodeB ID: 277.
Top1 neighbor: (same gNodeB), BRSRP -98.83 dBm, $\Delta$RSRP -4.16 dB.
Top2 neighbor: (different gNodeB), BRSRP -110.78 dBm, $\Delta$RSRP -16.11 dB.\\
\textbf{UE Global State}: Maximum UE speed during the test: 38.0 km/h.
Maximum UE-to-serving-cell distance during the test: 0.0856 km.\\
\textbf{Trajectory-Level Events}:
Total handover count during the test: 1.\\
\textbf{Serving Cell Engineering Parameters}:
Mechanical Azimuth: 60.
Mechanical Downtilt: 30.
Digital Tilt: 9.
Beam Scenario: DEFAULT.
Height: 30.1.\\
\textbf{Instruction}: Provide explanation with analyses and the final\_decision with ``root cause number'',\\
\\
\textbf{Explanation}: ``\textbf{[Step 1]:}  \texttt{Speed\_check}: False.   \texttt{Low RB\_check}: False.   \texttt{Handover\_check}: False.   \texttt{Distance\_check}: False.
\textbf{[Step 2]:} All checks False, candidate root causes that are related to ``\texttt{High\_speed}'', ``\texttt{Low RB}'', ``\texttt{Frequent\_handovers}'', and ``\texttt{Overshooting}'' are ruled out. Coverage loss at the cell edge due to aggressive Serving Cell Engineering Parameters (Serving cell Downtilt, angle and height parameters)''.\\
\textbf{Final\_decision}: ``M3''\}
\end{tcolorbox}
 \caption{A sample of SEKA-FT input/output.}
  \label{fig:example}
\end{figure*}

\subsection{Canonical Context Structuring for Stable RCA Fine-tuning}
\label{sec-stable}

In practical telecom environments, RCA tasks involve highly heterogeneous inputs such as troubleshooting tickets, user-plane tables, OSS system logs, etc.
Directly feeding these raw and loosely organized inputs into LLMs can introduce representational inconsistency and degrade fine-tuning stability.
Meanwhile, a well-resolved telecom RCA troubleshooting report typically requires standardized outputs, e.g., a clear root cause identification and structured reasoning steps.
Therefore, if similar evidence is presented in different formats across samples, the model may learn unstable reasoning trajectories even when the underlying diagnostic logic is the same.
To address this, we introduce canonical context structuring.
Given heterogeneous network inputs, this component normalizes the RCA context into consistent evidence blocks to:
\begin{itemize}
\item Explicitly separate evidence blocks, including user-plane performance indicators, control-plane signalling events, topology relationships, and configuration parameters.
As shown in Fig.~\ref{fig:example}, the input is organized into blocks such as \textbf{Bottleneck Snapshot} (e.g., \texttt{Downlink throughput}, \texttt{Serving SS-SINR}), \textbf{UE Global State} (e.g., \texttt{Maximum UE speed}), and \textbf{Trajectory-Level Events} (e.g., \texttt{Total handover count}).

\item Maintain consistent semantic slots across samples.
For instance, serving-cell metrics, neighbour-cell relationships, and mobility-related indicators are always presented in comparable structural positions.
This ensures that evidence such as \texttt{Scheduled RBs} or \texttt{Serving SS-RSRP} is interpreted within a fixed diagnostic context rather than as isolated numeric tokens.

\item Reduce non-essential noise, such as redundant log fragments or unrelated KPIs, while preserving diagnostically meaningful signals for root-cause inference.
\end{itemize}

Canonical structuring therefore provides a stable input interface between heterogeneous telecom telemetry and LLM-based diagnostic reasoning.
In the illustrative case, the structured presentation of \texttt{Mechanical Downtilt}, \texttt{Digital Tilt}, and radio measurements enables the model to associate aggressive antenna configuration with edge-coverage loss rather than misattributing the issue to scheduling or mobility.
This reduces output randomness, stabilizes diagnostic narratives, and promotes standardized RCA reporting behavior.

\begin{figure*}[t]
  \centering
  \includegraphics[width=0.98\textwidth]{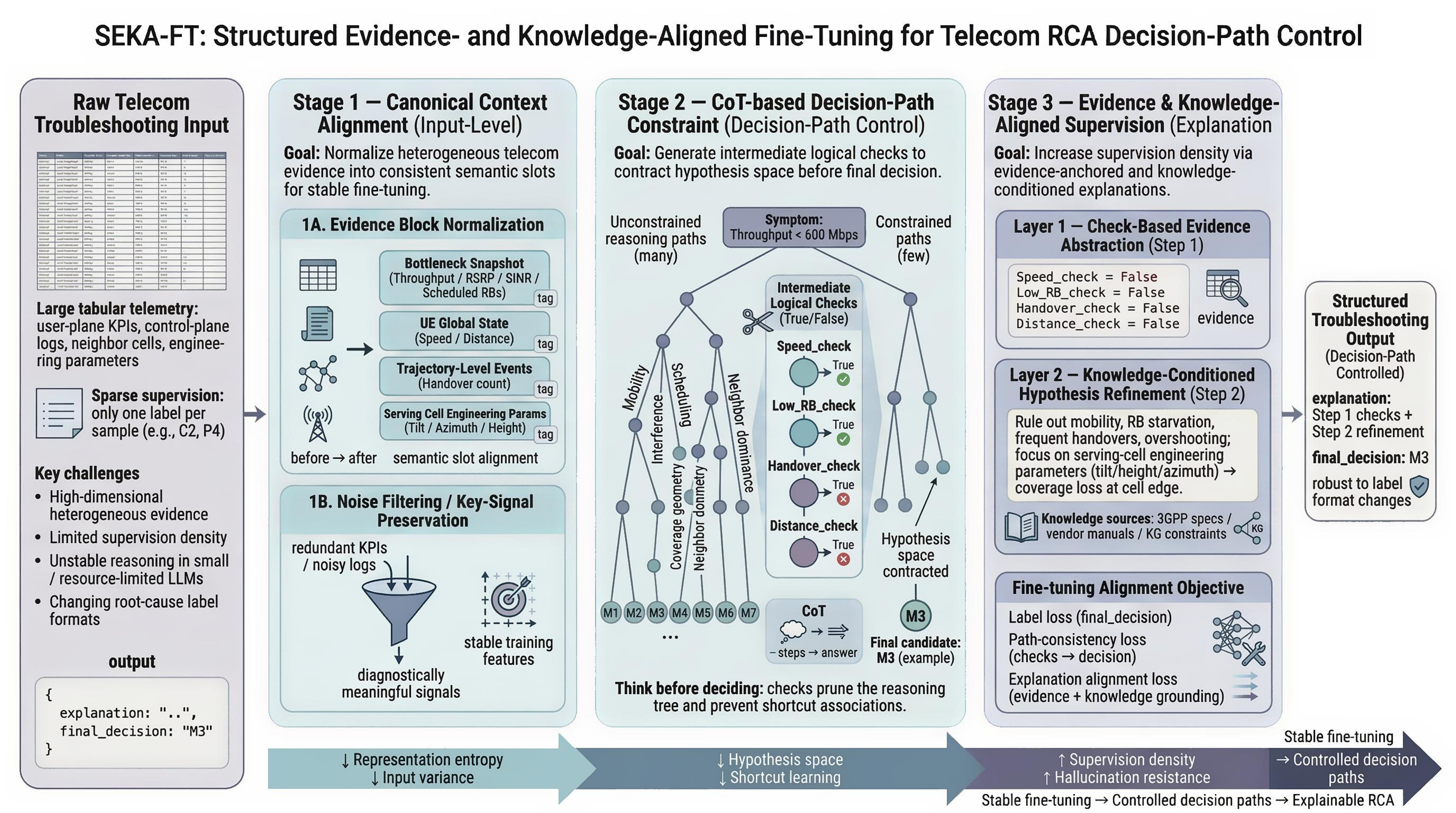} 
  \caption{Overview of the proposed structured evidence- and knowledge-aligned fine-tuning (SEKA-FT) framework.}
  \label{fig:seka}
  \vspace{-15pt}
\end{figure*}

\subsection{CoT-enabled Decision-Path Control for Reliable RCA}
\label{sec-cot}

Even with structured inputs, directly fine-tuning an LLM to predict final RCA labels remains insufficient.
A single categorical label provides sparse supervision and does not specify how the model should evaluate evidence before reaching a diagnosis.
As a result, the model may learn shortcut correlations between surface input patterns and root-cause labels rather than the intended troubleshooting logic.
For instance, in Fig.~\ref{fig:example}, the LLM may incorrectly associate moderate \texttt{Serving SS-RSRP} degradation with a generic coverage issue, or prematurely link the presence of \texttt{neighbour PCI} information to interference-related causes, without systematically verifying mobility, scheduling, and engineering parameters.
Such errors occur because the label-only objective does not constrain the reasoning trajectory used to produce the final answer.

To avoid implicit one-shot reasoning, SEKA-FT introduces CoT-enabled decision-path control.
Rather than directly mapping complex evidence to a final root-cause label such as ``M3'',
the model is supervised to first generate intermediate diagnostic assessments derived from the structured evidence.
In Fig.~\ref{fig:example}, this corresponds to structured checks such as whether \texttt{Maximum UE speed} exceeds a mobility threshold, whether \texttt{Scheduled RBs} fall below a resource sufficiency bound, whether \texttt{Total handover count} indicates mobility instability, or whether \texttt{Maximum UE-to-serving-cell distance} suggests potential coverage overshoot.
These checks convert high-dimensional telecom measurements into compact and verifiable evidence abstractions.
They also progressively prune unsupported hypotheses before the final RCA decision is generated.
In some cases, a decisive abnormality detected at this stage, such as excessive mobility speed, may already be sufficient to explain the throughput degradation without further hypothesis exploration.
Therefore, CoT in SEKA-FT is not used merely as a prompting trick, but as a structured supervision mechanism embedded into fine-tuning.
It transforms RCA from direct classification into controlled hypothesis refinement, improving both diagnostic reliability and interpretability.

\subsection{Evidence- and Knowledge-Anchored Explanation Design for Hallucination Mitigation}
\label{sec-evidence}

While CoT-enabled reasoning improves the decision trajectory, effective learning of diagnostic behavior still requires sufficiently rich supervision signals.
In many telecom RCA datasets, the available supervision is only a compact root-cause label, which is too sparse for teaching the model how to select evidence, eliminate unsupported hypotheses, and justify the final decision.
To enable the model to learn diagnostic logic rather than memorize label associations, we introduce evidence- and knowledge-anchored explanation design as a core component of SEKA-FT.
The key idea is to use explanations as structured supervision carriers, so that the training target encodes not only the final answer but also the evidence path leading to that answer.
As shown in Fig.~\ref{fig:example}, the designed explanation follows a two-step evidence-based structure.

Specifically, as introduced in Section~\ref{sec-cot}, \textbf{Step 1} extracts simplified diagnostic evidence from structured inputs in the form of interpretable checks.
Rather than directly predicting the final root cause, the model first evaluates core dimensions of telecom troubleshooting, such as mobility behavior (\texttt{Maximum UE speed}) and scheduling sufficiency (\texttt{Scheduled RBs}).
These check results explicitly indicate which causal dimensions are supported or contradicted by the observed telemetry.
Therefore, Step 1 provides a controllable evidence basis for subsequent hypothesis refinement.

Then, \textbf{Step 2} performs knowledge-guided decision refinement conditioned on the outcomes of Step 1.
In the illustrative case, all checks (\texttt{Speed\_check}, \texttt{Low RB\_check}, \texttt{Handover\_check}, and \texttt{Distance\_check}) are False.
This evidence eliminates mobility-driven degradation, resource starvation, frequent handovers, and excessive coverage distance as primary causes.
After these hypotheses are ruled out, the model narrows its reasoning focus to serving-cell engineering parameters, including \texttt{Mechanical Downtilt}, \texttt{Digital Tilt}, and antenna height.
Thus, the model does not reconsider all possible root causes from scratch, but refines its hypothesis space according to validated evidence and domain knowledge.
This structured transition from evidence extraction to knowledge-conditioned hypothesis refinement embodies the core logic of the proposed multi-stage reasoning design.
By forcing explanations to remain anchored to observable indicators, SEKA-FT reduces hallucinated reasoning and improves the traceability of RCA outputs.

\subsection{Integrated SEKA-FT Framework and Alignment Objective}
\label{sec-full}
Fig.~\ref{fig:seka} summarizes the overall procedure of the proposed SEKA-FT framework.
The pipeline starts from raw telecom troubleshooting inputs, which contain heterogeneous evidence such as user-plane KPIs, control-plane logs, neighbor-cell relationships, and serving-cell engineering parameters.
Since these inputs are usually paired with sparse supervision signals, such as a single root-cause label ``M3'', directly fine-tuning a small LLM may lead to unstable or shortcut-based reasoning.

To address this issue, Stage 1 performs canonical context alignment at the input level.
Heterogeneous evidence is normalized into consistent semantic slots, including \texttt{bottleneck snapshots}, \texttt{UE global states}, \texttt{trajectory-level events}, and \texttt{engineering parameters}.
This stage stabilizes the input feature space and provides a consistent evidence interface for downstream reasoning.
Stage 2 then introduces CoT-based decision-path control.
Instead of directly mapping structured evidence to a final root-cause label, the model first generates intermediate diagnostic checks, such as mobility, scheduling sufficiency, handover instability, and coverage distance.
These checks prune unsupported hypotheses and reduce the effective RCA decision space before the final prediction is produced.
As visualized in the tree-pruning diagram, the reasoning process is progressively contracted from unconstrained label selection into a smaller set of plausible candidates.

Stage 3 further integrates evidence- and knowledge-aligned explanation supervision into the structured target.
The supervision signal is expanded from a single categorical label to a multi-layer reasoning sequence: Layer 1 encodes check-based evidence abstraction derived from structured telemetry, while Layer 2 performs knowledge-conditioned hypothesis refinement grounded in domain constraints.
This design ensures that diagnostic conclusions are not only accurate at the label level, but also traceable to validated telecom evidence and engineering knowledge.

During fine-tuning, SEKA-FT optimizes a token-level causal language modeling objective over the structured target sequence, with prompt tokens masked out.
\textcolor{black}{The structured target jointly encodes the intermediate diagnostic path, evidence-grounded hypothesis refinement, and final RCA decision, thereby preserving the dependency from observed telecom evidence to diagnostic reasoning and ultimately to root-cause identification.}
\textcolor{black}{Rather than supervising these elements independently, SEKA-FT couples them within a unified evidence-to-path-to-decision learning structure, allowing each diagnostic conclusion to be progressively grounded in preceding evidence and domain constraints.}
\textcolor{black}{Therefore, through this joint alignment, SEKA-FT transforms fine-tuning from direct label supervision into structured diagnostic reasoning supervision, improving supervision density, hallucination resistance, reasoning stability, and the traceability of RCA outputs.}

\section{Case Study: Performance Evaluation in 5G Throughput Degradation}

\subsection{Experiment Settings}
We evaluate SEKA-FT on two 5G RCA datasets:
\begin{enumerate}[label=\arabic*., leftmargin=*]
\item The primary evaluation is conducted on \textbf{TeleLogs}~\cite{sana2025reasoning}, a telecom-specific benchmark associated with the GSMA LLM benchmarks initiative. Each sample consists of structured user-plane KPIs, mobility statistics, serving-cell engineering parameters, and a corresponding root-cause label.
\item We further evaluate SEKA-FT on \textbf{TelecomTS}~\cite{feng2025telecomts}, a 5G observability dataset derived from a lab-deployed testbed with high-resolution KPI records collected from both the base station and user device under live application traffic.
\end{enumerate}
\blue{These two benchmarks provide complementary evaluation settings: TeleLogs emphasizes joint reasoning over drive-test measurements, neighbor-cell relations, and engineering configurations, whereas TelecomTS introduces testbed-based multi-channel temporal KPI observations under diverse network conditions.}

For SEKA-FT, the structured supervision target is constructed from the original RCA label, benchmark-defined diagnostic checks, and automatically generated evidence-grounded explanations. During fine-tuning, the target sequence contains the intermediate diagnostic path, the hypothesis-refinement explanation, and the final RCA decision. 
The primary base model is Qwen2.5-1.5B-Instruct, and we compare SEKA-FT against the following baselines using supervised fine-tuning (SFT) where applicable:
1) \textbf{Regular ICL} (in-context learning, no SFT);
2) \textbf{LSTM} (non-LLM sequence-classification baseline);
3)  \textbf{Vanilla SFT} (using the raw input/output from TeleLogs);
4) \textbf{SFT + Structured Input} (as introduced in Section~IV-A);
5) \textbf{SFT + Explanation} (as introduced in Sections~IV-B and~C);
6) \textbf{SEKA-FT} (full model as shown in Section~IV-D).
To study the effect of model scaling, we additionally evaluate Qwen2.5-7B-Instruct and Qwen3-32B.
LoRA is applied to the last four Transformer blocks for parameter-efficient fine-tuning; Training uses causal LM loss with prompt tokens masked out; we optimize with AdamW (learning rate 1e‑5), apply gradient clipping and train for 12 epochs with batch size 2 and max sequence length 2048; we use mixed precision with bf16. The experiment is implemented on Nvidia RTX 6000 Ada.

\blue{All reported results are averaged over ten independent runs with different random seeds and presented with 95\% confidence intervals, calculated using Student's $t$-interval based on the sample mean and standard deviation across the ten runs. In addition, we conduct paired statistical significance tests using predictions on the same held-out samples. Exact McNemar's test is used for Accuracy, while a paired permutation test is used for Macro-F1. When multiple baselines are considered, the resulting $p$-values are adjusted using the Holm--Bonferroni procedure.}

\begin{figure*}[!t]
\centering
\subfigure[\textbf{Accuracy and Macro-F1 results comparison.}]{
\includegraphics[width=5.5cm,height=4.2cm]{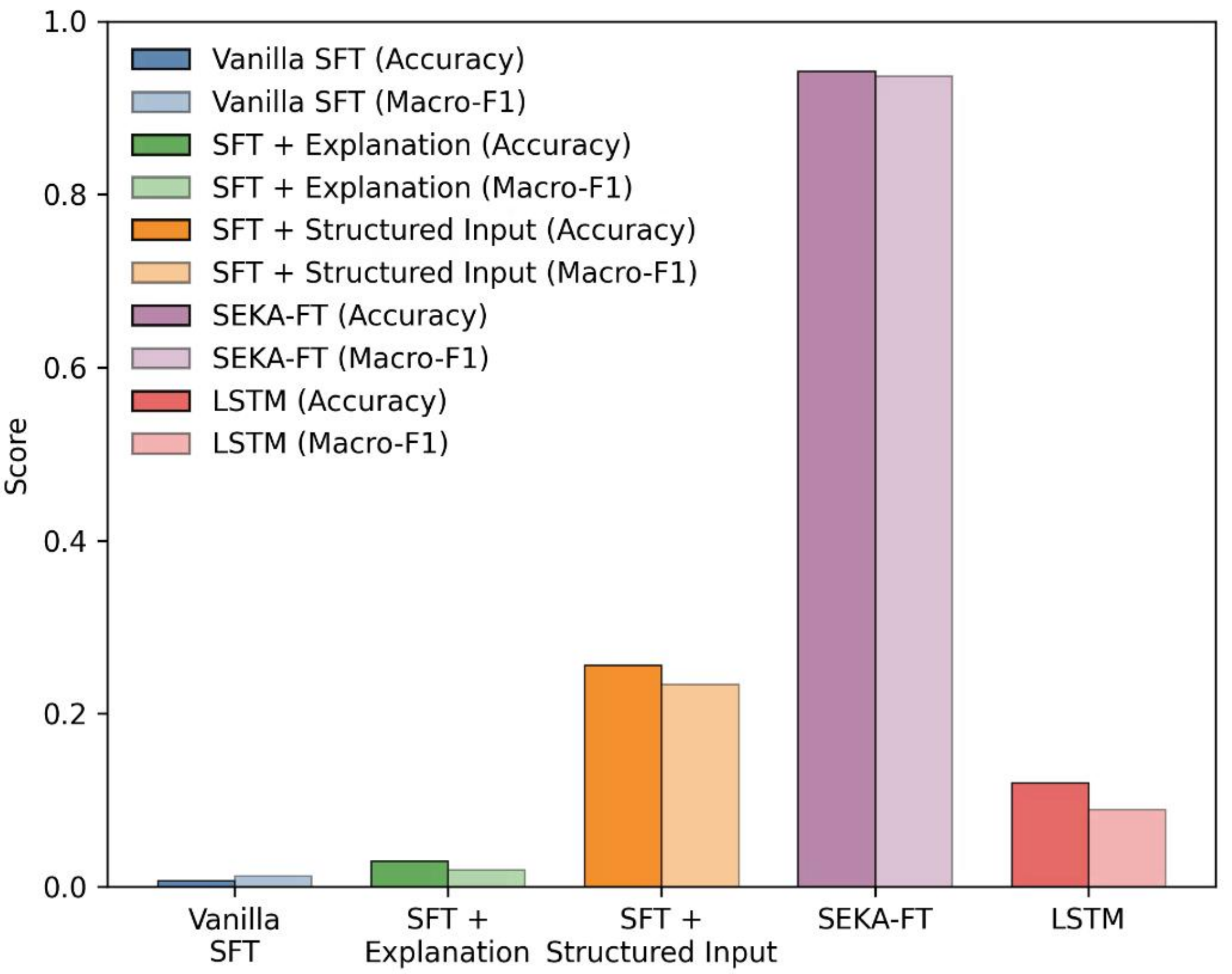} 
\label{f-r1}
}
\hfill
\subfigure[Convergence dynamics comparison.]{
\includegraphics[width=5.5cm,height=4.2cm]{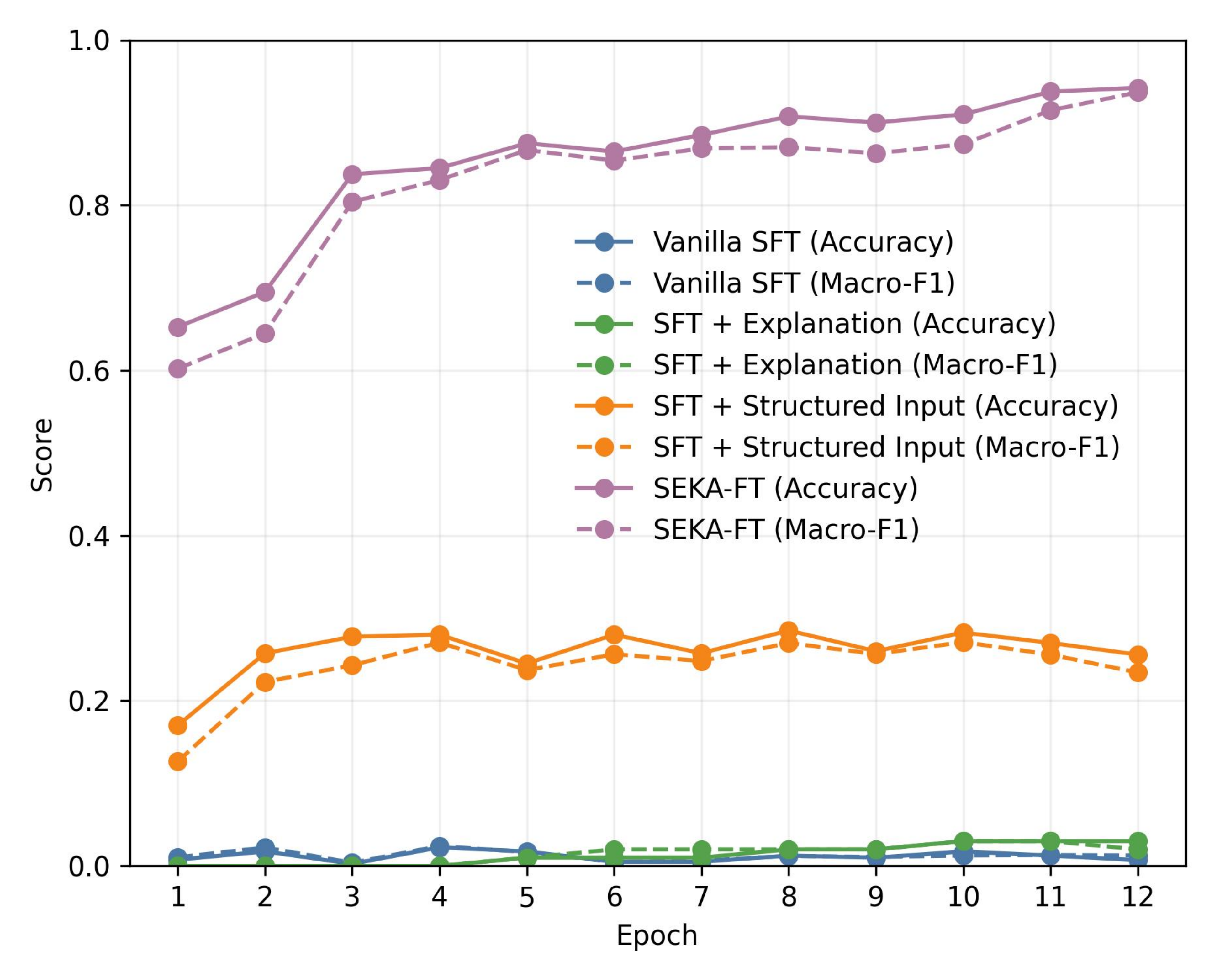} \label{f-r2}
}
\hfill
\subfigure[Comparison between ICL performance of Qwen2.5-1.5B, 7B, and 32B.]{
\includegraphics[width=5.5cm,height=4.2cm]{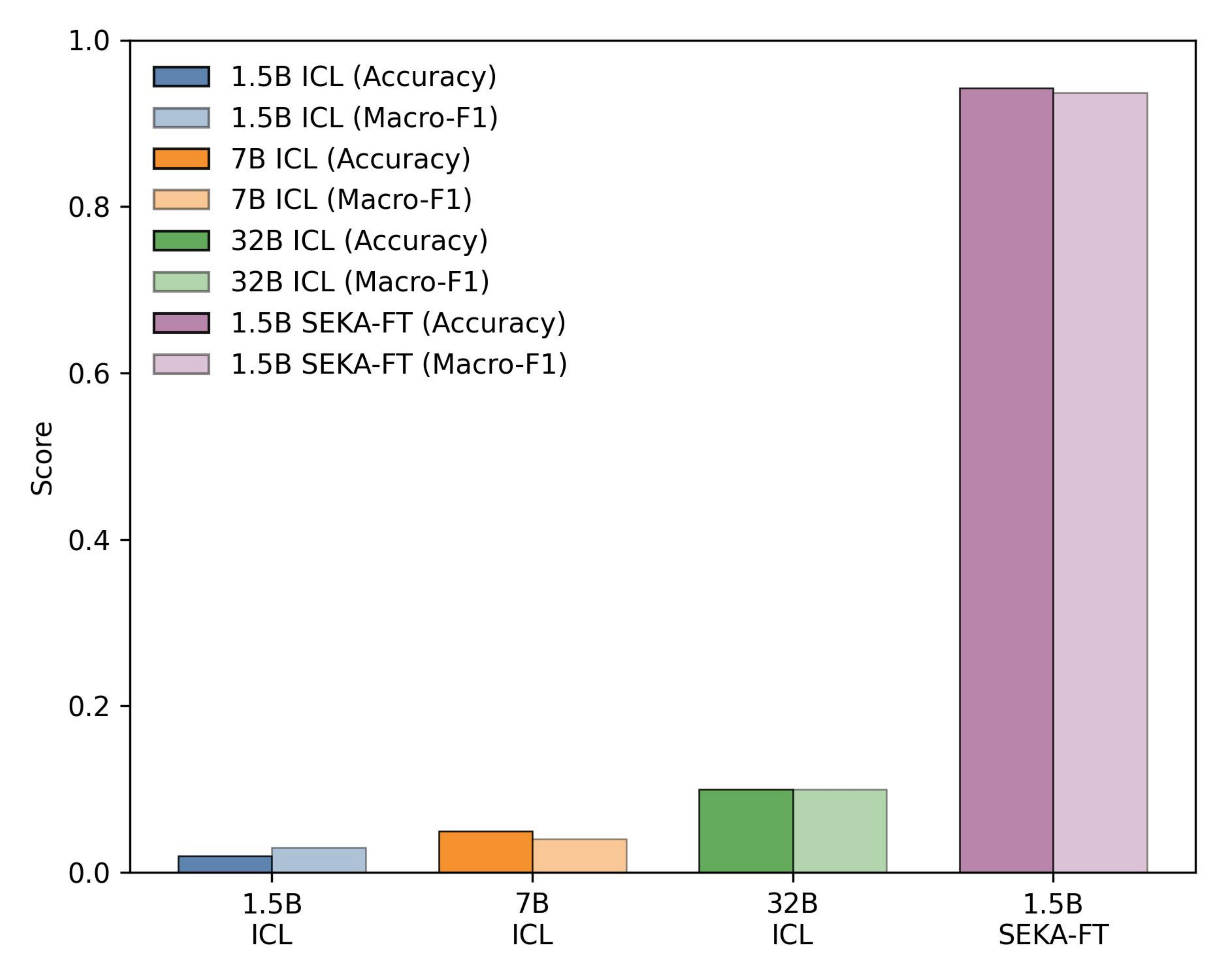} 
\label{f-r3}
}
\\
\subfigure[RCA decision path consistency accuracy comparison.]{
\includegraphics[width=5.5cm,height=4.2cm]{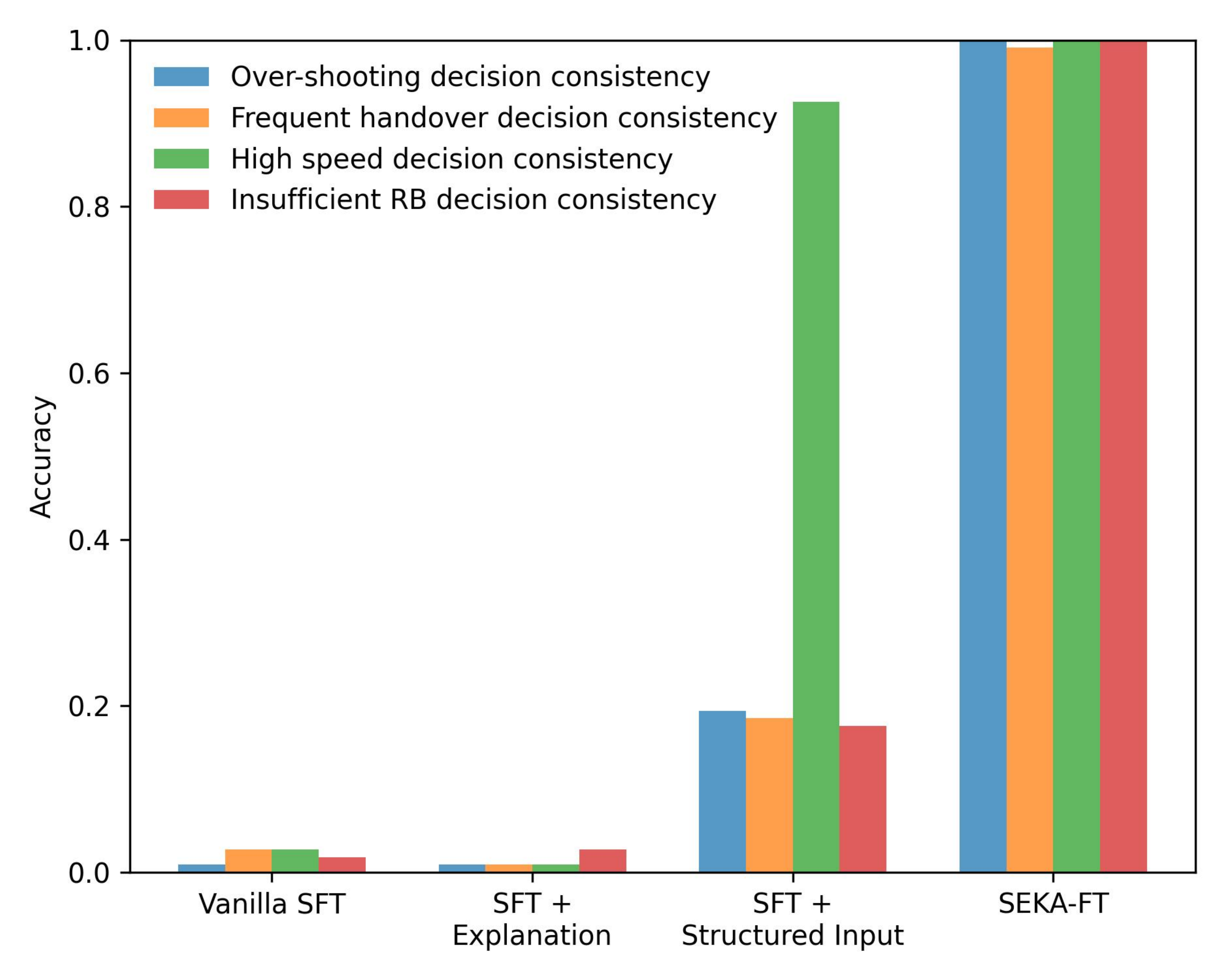} 
\label{f-r4}
}
\hfill
\subfigure[\textbf{Accuracy analyses of all root causes.}]{
\includegraphics[width=5.5cm,height=4.2cm]{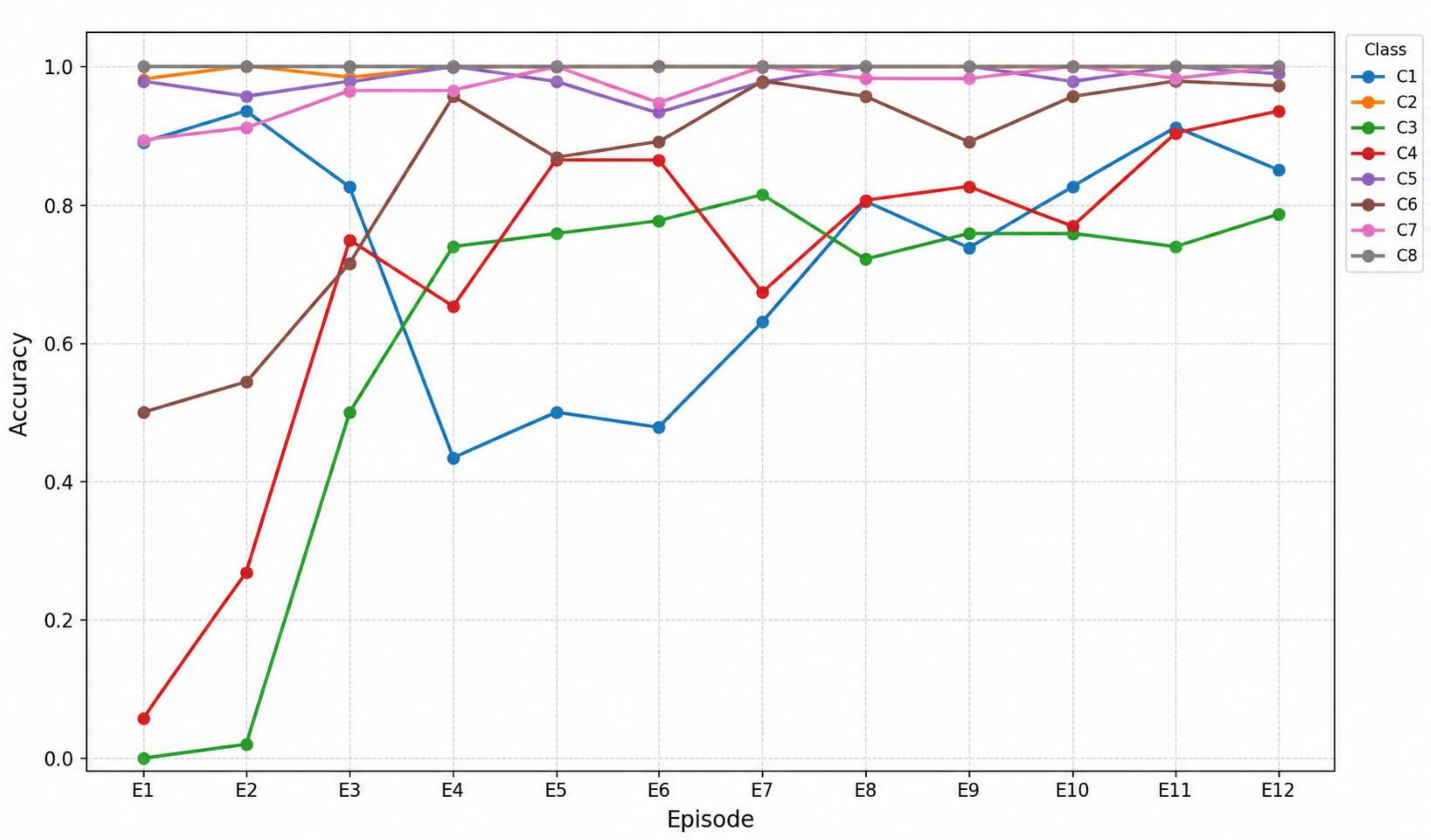} 
\label{f-r5}
}
\hfill
\subfigure[\textbf{RCA candidate-space reduction via sequential diagnostic checks.}]{
\includegraphics[width=5.5cm,height=4.2cm]{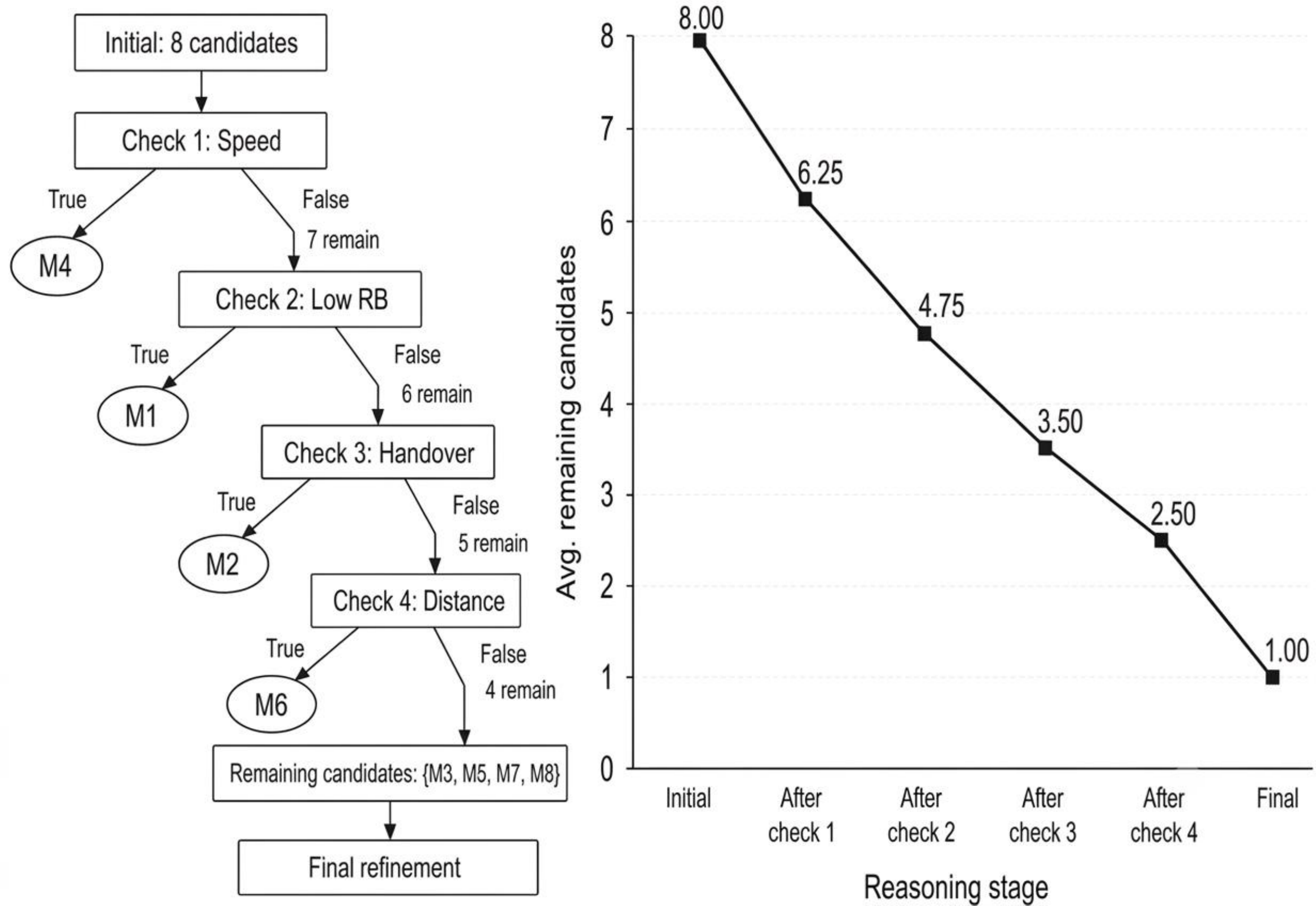} 
\label{f-r6}
}
\caption{Experiment results and comparisons.}
\label{fig_all}
\vspace{-15pt}
\end{figure*}

\subsection{Experiment Results on TeleLogs}

We first evaluate the proposed framework on TeleLogs, which serves as the primary benchmark in this study. All reported results are averaged over multiple runs with different random seeds and presented with 95\% confidence intervals.

Fig.~\ref{f-r1} compares the overall Accuracy and Macro-F1 of different training strategies. Vanilla SFT performs poorly, achieving only $0.007 \pm 0.003$ Accuracy and $0.013 \pm 0.004$ Macro-F1, while SFT + Explanation reaches only $0.030 \pm 0.005$ and $0.020 \pm 0.004$, respectively. These results show that neither direct fine-tuning on raw inputs nor explanation supervision alone is sufficient. SFT + Structured Input improves performance to $0.256 \pm 0.018$ Accuracy and $0.234 \pm 0.016$ Macro-F1, confirming the benefit of canonical context structuring, but remains substantially below SEKA-FT. SEKA-FT achieves the best performance, reaching $0.942 \pm 0.006$ Accuracy and $0.937 \pm 0.007$ Macro-F1. The LSTM baseline achieves $0.132 \pm 0.012$ Accuracy and $0.090 \pm 0.011$ Macro-F1, outperforming Vanilla SFT but remaining considerably weaker than the structured LLM-based approaches.

Fig.~\ref{f-r2} shows the convergence dynamics across training epochs. Vanilla SFT and SFT + Explanation remain close to random-level performance, while SFT + Structured Input converges more stably but saturates at a substantially lower level. In contrast, SEKA-FT improves rapidly and maintains consistently high Accuracy and Macro-F1, indicating that structured supervision facilitates more effective learning of both final labels and intermediate diagnostic logic.

Fig.~\ref{f-r3} evaluates model scaling under regular in-context learning. Although increasing the model size improves ICL performance to some extent, the absolute performance remains low. Qwen3-32B reaches only $0.126 \pm 0.010$ Accuracy and $0.109 \pm 0.009$ Macro-F1. In comparison, SEKA-FT with the lightweight 1.5B model substantially outperforms all ICL baselines, showing that structured fine-tuning is more effective than simply increasing model size for this RCA task.

Fig.~\ref{f-r4} evaluates decision-path consistency across representative diagnostic checks, including overshooting coverage, frequent handovers, high mobility speed, and insufficient resource blocks. These checks are deterministic evidence abstractions derived from the original RCA definitions rather than thresholds tuned on test results; for example, the high-speed and insufficient-RB checks follow the definitions of vehicle speed exceeding 40 km/h and average scheduled RBs below 160, respectively. The explanation targets are automatically constructed from the ground-truth label, deterministic check results, and domain-rule-based hypothesis refinement templates. Vanilla SFT and SFT + Explanation exhibit very low decision-path consistency, while SFT + Structured Input improves some explicit checks but remains weak on several others. In contrast, SEKA-FT achieves near-perfect consistency across all evaluated checks, demonstrating improved logical stability of intermediate diagnostic reasoning in addition to final RCA accuracy.

Fig.~\ref{f-r5} analyzes the difficulty of individual RCA categories. C2 and C8 are almost perfectly classified, while C5 and C7 also achieve consistently high accuracy, indicating relatively distinctive evidence patterns. Most remaining errors occur among coverage- and neighbor-cell-related categories, where excessive downtilt, stronger neighboring-cell selection, and overlapping coverage can produce similar RSRP/SINR and neighbor-cell patterns. Thus, the residual errors are concentrated in telecom-meaningful ambiguous cases rather than being randomly distributed.

Finally, Fig.~\ref{f-r6} illustrates how sequential diagnostic checks reduce the effective RCA decision space. The model first evaluates four intermediate checks; a positive check can directly identify the corresponding root cause, while a negative result prunes that hypothesis. Consequently, the original 8-way decision space is progressively reduced, with only four candidates remaining when none of the four checks is triggered. Under a uniform root-cause distribution, the average candidate-set size decreases from 8 to 6.25, 4.75, 3.5, and 2.5 after the first through fourth checks, respectively. This demonstrates how structured intermediate checks reduce decision difficulty before the final RCA refinement.

\begin{table}[!t]
\centering
\begingroup
\bfseries\boldmath
\caption{Comprehensive experimental results on the TeleLogs and TelecomTS datasets.}
\label{tab:combined_telelogs_telecomts_results}
\small
\setlength{\tabcolsep}{5.5pt}
\renewcommand{\arraystretch}{1.15}
\resizebox{\columnwidth}{!}{%
\begin{tabular}{lcccc}
\toprule
\multirow{2}{*}{Method}
& \multicolumn{2}{c}{TeleLogs}
& \multicolumn{2}{c}{TelecomTS} \\
\cmidrule(lr){2-3}
\cmidrule(lr){4-5}
& Accuracy
& Macro-F1
& Accuracy
& Macro-F1 \\
\midrule
ICL 1.5B
& $0.021 \pm 0.004$
& $0.018 \pm 0.003$
& $0.033 \pm 0.006$
& $0.031 \pm 0.005$ \\

ICL 7B
& $0.053 \pm 0.006$
& $0.043 \pm 0.005$
& $0.041 \pm 0.005$
& $0.028 \pm 0.003$ \\

ICL 32B
& $0.126 \pm 0.010$
& $0.109 \pm 0.009$
& $0.061 \pm 0.005$
& $0.053 \pm 0.004$ \\

LSTM
& $0.132 \pm 0.012$
& $0.090 \pm 0.011$
& $0.115 \pm 0.010$
& $0.080 \pm 0.010$ \\

Vanilla SFT
& $0.007 \pm 0.003$
& $0.013 \pm 0.004$
& $0.050 \pm 0.021$
& $0.052 \pm 0.016$ \\

\makecell[l]{SFT+\\Struct Input}
& $0.256 \pm 0.018$
& $0.234 \pm 0.016$
& $0.132 \pm 0.009$
& $0.138 \pm 0.009$ \\

\makecell[l]{SFT+\\Explanation}
& $0.030 \pm 0.005$
& $0.020 \pm 0.004$
& $0.047 \pm 0.008$
& $0.041 \pm 0.008$ \\

SEKA-FT
& $\mathbf{0.942 \pm 0.006}$
& $\mathbf{0.937 \pm 0.007}$
& $\mathbf{0.647 \pm 0.004}$
& $\mathbf{0.615 \pm 0.005}$ \\
\bottomrule
\end{tabular}%
}
\endgroup
\end{table}

\subsection{Experiment Results on TelecomTS}
We further evaluate SEKA-FT on TelecomTS as an independent cross-dataset validation. This experiment is used to examine whether the proposed structured reasoning design generalizes beyond the TeleLogs-specific input format.

As shown in Table~\ref{tab:combined_telelogs_telecomts_results}, the overall trend on TelecomTS is highly consistent with the TeleLogs results. Regular ICL remains ineffective across all model sizes. ICL with 1.5B, 7B, and 32B models achieves only $0.033 \pm 0.006$, $0.041 \pm 0.005$, and $0.061 \pm 0.005$ Accuracy, respectively.
The non-LLM LSTM baseline achieves $0.115 \pm 0.010$ Accuracy and $0.080 \pm 0.010$ Macro-F1, which is slightly better than most ICL and raw SFT baselines, but still far below SEKA-FT. Vanilla SFT also performs poorly, with $0.050 \pm 0.021$ Accuracy. Similarly, SFT + Explanation on raw numerical inputs reaches only $0.047 \pm 0.008$ Accuracy, confirming that explanation supervision alone is insufficient without structured evidence alignment.
In contrast, SFT + Structured Input improves the result to $0.132 \pm 0.009$ Accuracy and $0.138 \pm 0.009$ Macro-F1, showing that canonical evidence organization is also beneficial on TelecomTS. More importantly, SEKA-FT achieves the best performance, reaching $0.647 \pm 0.004$ Accuracy and $0.615 \pm 0.005$ Macro-F1. Compared with the strongest non-SEKA baseline, SFT + Structured Input, SEKA-FT improves Accuracy by $0.515$ and Macro-F1 by $0.477$.

\blue{
To further assess whether the observed performance gains are statistically significant, we conduct paired comparisons between SEKA-FT and the strongest competing baseline, SFT + Structured Input, using predictions on the same held-out samples.
Specifically, we apply the exact McNemar's test to Accuracy and a paired permutation test to Macro-F1.
On TeleLogs, SEKA-FT significantly outperforms SFT + Structured Input in both Accuracy and Macro-F1 ($p<0.001$ for both metrics).
The same conclusion is observed on TelecomTS, where the improvements in both Accuracy and Macro-F1 are also statistically significant ($p<0.001$ for both metrics).
These results confirm that the substantial performance gains of SEKA-FT over the strongest competing baseline are consistently supported by paired statistical significance analysis across both datasets.}

\blue{In summary, the ranking of these methods is consistent across the two datasets: raw SFT and ICL are ineffective, conventional approaches such as LSTM provide only limited improvement, structured input helps but remains insufficient, and the full SEKA-FT framework achieves the strongest performance. This cross-dataset consistency demonstrates that the effectiveness of SEKA-FT extends beyond the TeleLogs-specific data representation and remains robust across distinct 5G RCA and observability settings. These results establish cross-dataset robustness within the evaluated 5G RAN/observability scope. Extending the proposed structured reasoning design to core-network, transport-network, alarm-topology, multi-domain, and real operator incident RCA represents an important direction for future evaluation.}

\section{Conclusion}

RCA plays a critical role in modern telecom networks, and this paper studied the application of LLMs for telecom RCA. We proposed a structured reasoning framework that aligns diagnostic reasoning with telecom-specific evidence and decision paths. Experimental results demonstrate that the proposed approach significantly improves diagnostic accuracy and decision consistency compared with conventional SFT and ICL baselines. These findings highlight the importance of structured reasoning design for reliable LLM-enabled RCA in next-generation telecom networks.
\blue{As future work, we will further evaluate and extend the proposed framework in more diverse and realistic telecom RCA scenarios, including core- and transport-network faults as well as previously unseen and real operator incidents.}

\normalem
\bibliographystyle{IEEEtran}
\bibliography{Reference}

\begin{IEEEbiography}
{Hao Zhou} is a Research Engineer at Samsung Research America, focusing on the intersection of large language models (LLMs) and 6G network intelligence. He received his Ph.D. in Electrical and Computer Engineering from the University of Ottawa in 2023 and was a Postdoctoral Researcher at McGill University. Dr. Zhou’s research spans reinforcement learning, LLMs for telecom, and AI-driven network optimization. His work has been recognized with the IEEE ICC 2023 Best Paper Award, the IEEE ComSoc CSIM TC Best Journal Paper Award, and the Best Doctoral Thesis Award from the University of Ottawa. He has also led multiple industry-academia collaborations with Ericsson Canada and Samsung Research America.
\end{IEEEbiography}

\begin{IEEEbiography}
{Mandar N. Kulkarni} received Ph.D. in Electrical Engineering from the University of Texas at Austin in 2018 and B. Tech in Electronics and Communications Engineering from the Indian Institute of Technology (IIT) Guwahati in 2013. He works in the area of network automation as a Sr. Staff Engineer at Samsung Research America, where he received President's award in 2018 and 2019. He received the President of India Gold Medal in 2013, George J. Heuer, Jr. Fellowship at UT Austin in 2014, WNCG student leadership award in 2018, IEEE Leonard G. Abraham prize in 2018. He has 18 publications and over 20 filed patents.
\end{IEEEbiography}

\begin{IEEEbiography}
{Hao Chen} received the B.S. and M.S. degrees in information engineering from Xi’an Jiaotong University, in 2010 and 2013, respectively, and the Ph.D. degree in electrical engineering from the University of Kansas, Lawrence, KS, USA, in 2017. He is currently a Director with Samsung Research America, where he is working on AI/agentic AI for wireless communication, wireless sensing and localization, network automation and localization.
\end{IEEEbiography}

\begin{IEEEbiography}
{Yan Xin} received the Ph.D. degree in electrical engineering from the University of Minnesota, Minneapolis, MN, USA. From 2004 to 2008, he was an Assistant Professor with the Department of Electrical and Computer Engineering, National University of Singapore. He is currently a Senior Research Director at the Standards and Mobility Innovation Laboratory, Samsung Research America. His research interests include network automation, AI-RAN PHY/MAC optimization, MIMO communications, and LTE/5G-NR/6G. He received the 2004 IEEE Marconi Prize Paper Award in wireless communications from the IEEE Communications Society.
\end{IEEEbiography}

\begin{IEEEbiography}
{Jianzhong (Charlie) Zhang} is an EVP at Samsung Research America, where he leads research, prototyping, and standardization for 5G/6G and other wireless systems. He is also a Corporate VP and head of the  global 6G team at Samsung Research. He is currently serving as the ATIS North America Next-G Alliance Full Member Group Vice Chair. He was the Board Chair of the FiRa Consortium  from May 2019 to May 2023, and the Vice Chairman of the 3GPP RAN1 working group from 2009 to 2013, where he led development of LTE and LTE-Advanced technologies.  He received his Ph.D. degree from the University of Wisconsin, Madison.  Dr. Zhang is a Fellow of IEEE  and received the 2024 IEEE ComSoc Industrial Innovation Award.
\end{IEEEbiography}

\end{document}